\documentclass[conference]{IEEEtran}
\usepackage{cite}
\usepackage{amsmath,amssymb,amsfonts}
\usepackage{algorithmic}
\usepackage{graphicx}
\usepackage{textcomp}
\usepackage{xcolor}
\usepackage[hyphens]{url}
\usepackage{hyperref}
\usepackage{booktabs}
\usepackage{array}
\usepackage{multirow}

\def\BibTeX{{\rm B\kern-.05em{\sc i\kern-.025em b}\kern-.08em
    T\kern-.1667em\lower.7ex\hbox{E}\kern-.125emX}}

\title{From Human Guidance to Autonomy:\\Agent Skill System for End-to-End LLM Deployment on Spatial NPUs}

\IEEEoverridecommandlockouts  

\author{%
\textbf{Jiajie Li}\thanks{Jiajie Li performed this work at AMD.}\quad
\textbf{Erwei Wang}$^{\ast}$\quad
\textbf{Zhiru Zhang}\quad
\textbf{Samuel Bayliss}$^{\ast}$\\
Cornell University\quad
$^{\ast}$AMD
\\
\footnotesize{\texttt{{\{\href{mailto:jl4257@cornell.edu}{jl4257},\,\href{mailto:zhiruz@cornell.edu}{zhiruz}\}@cornell.edu}}}\,,\footnotesize{\texttt{\{\href{mailto:erwei.wang@amd.com}{erwei.wang},\,\href{mailto:samuel.bayliss@amd.com}{samuel.bayliss}\}@amd.com}}
}
\usepackage{fancyhdr}
\fancypagestyle{firstpage}{
  \fancyhf{} 
  \chead{\small \textit{Accepted to the Machine Learning for Architecture and Systems Workshop (\href{https://sites.google.com/corp/view/mlarchsys}{MLArchSys}), co-located with \href{https://iscaconf.org/isca2026/}{ISCA 2026}.}}
  \cfoot{\thepage} 
}

\begin{document}
\maketitle

\thispagestyle{firstpage} 
\pagestyle{plain}


\begin{abstract}
Spatial neural processing units (NPUs) provide an energy-efficient
platform for edge LLM inference, but efficiently deploying an LLM end-to-end on such hardware remains labor-intensive. 
Although AI coding agents have begun to lower this cost, existing studies have largely focused on single-kernel optimization rather than end-to-end LLM deployment on resource-constrained spatial NPUs.

We present a two-stage methodology, instantiated on the
AMD XDNA\texttrademark~2 NPU, that progresses from
human-guided development to agent autonomy.
In the first stage, we develop a reference deployment of Llama-3.2-1B through human-guided agent assistance. 
The resulting implementation achieves a speedup of 2.2$\times$ on prefill and 4.0$\times$ on decode over the hand-optimized baseline, with the
optimization trajectory and its lessons recorded as structured documentation throughout. In the second stage, we distill the documentation into an agent skill system consisting of eight phases, orchestrating the optimization and debugging skill sets, with numerical correctness strictly enforced at each phase.

Using our agent skill system, we autonomously deploy eight
additional decoder-only LLMs (Llama-3.2-3B, SmolLM2-1.7B,
Qwen2.5-\{0.5B, 1.5B, 3B\}, Qwen3-\{0.6B, 1.7B, 4B\}) end-to-end on the
AMD XDNA~2 NPU using the open-source compiler stack. 
To our knowledge, these models have not previously been deployed on AMD NPUs via any open-source software stack. Each deployment completes in 0.5--4 hours of agent wall time
with almost no human guidance, and passes the numerical-correctness gates, demonstrating functional generalization to previously unencountered LLMs. Three of the eight match or exceed
the sustained performance of our Llama-3.2-1B reference deployment, suggesting that the resulting implementations can be competitive without additional model-specific human engineering.
\end{abstract}

\section{Introduction}

Large language models are increasingly deployed at the edge for
lower latency, stronger privacy, and offline operation. These settings impose strict power and thermal constraints, motivating the use of accelerators with higher energy efficiency than general-purpose CPUs or GPUs. Spatial neural processing
units (NPUs) have emerged as a key class of accelerator for this
regime.

Spatial NPUs achieve efficiency by exposing explicit hardware management to the software stack.
Unlike processors with implicit cache
hierarchies, spatial NPUs expose distributed on-chip memories,
explicit data-movement scheduling, and tile-level kernel placement
directly to the programmer. Deploying an LLM end-to-end onto NPUs with competitive
performance typically takes substantial expert engineering to tackle
all these challenges.

Agentic systems for accelerator
programming~\cite{kernelagent,npueval,autocomp,kernelevolve,astra,accelopt,geak}
reduce engineering effort by automating kernel generation.
However, no existing research primarily targets end-to-end LLM deployment on a resource-constrained spatial NPU.
They also do not encode the full deployment workflow as composable Agent Skills~\cite{anthropicskills} that can be invoked autonomously by a coding agent.
We address both gaps for end-to-end LLM deployment on the
AMD XDNA\texttrademark~2 NPU~\cite{NPUoverview}.

Our approach consists of two stages, embodying a transition \textbf{from
human guidance to agent autonomy}. In the first stage, we construct a reference Llama-3.2-1B
on NPU with human guidance and agent assistance, and document the
optimization trajectory and its lessons throughout. In the second stage, we distill the
documentation into a reusable agent skill system that supports the
autonomous deployment of additional LLMs. The paper
makes three novel contributions:

\begin{itemize}
  \item An end-to-end Llama-3.2-1B deployment on the AMD XDNA~2 NPU
        (\S\ref{sec:bootstrap}) that achieves a speedup of
        2.2$\times$ on prefill and 4.0$\times$ on decode over the
        hand-optimized baseline. The optimization trajectory and
        its lessons are maintained as documents throughout, forming
        the foundation of the skill system.
  \item An agent skill system for end-to-end LLM deployment on
        NPUs (\S\ref{sec:skills}), comprising an eight-phase
        skill chain with strict numerical correctness gates, sets
        of optimization and debug skills that auto-trigger on common
        patterns, and an independent evaluator agent that re-runs
        every gate to prevent the generator from bypassing it.
  \item First open-source end-to-end deployment of eight additional
        LLMs on the AMD XDNA~2 NPU (\S\ref{sec:eval}): Llama-3.2-3B, SmolLM2-1.7B,
        Qwen2.5-\{0.5B, 1.5B, 3B\}, and Qwen3-\{0.6B, 1.7B, 4B\}.
        Each deploys autonomously in 0.5--4 hours of agent wall time, with three
        of them matching or exceeding the sustained performance of our
        Llama-3.2-1B reference.
\end{itemize}

\section{Background}
\label{sec:background}

\noindent\textbf{Target hardware.} We target the AMD XDNA~2
NPU in Ryzen\texttrademark~AI 300/400 Series processors.
Compute tiles are arranged in a 4$\times$8 array, each with
a 64\,KB L1 scratchpad. Each column also includes a
512\,KB memory tile (L2) shared by its four compute tiles, plus
a shim tile bridging to host DDR. Compute tiles communicate
via configurable streaming interconnects and cascade
connections to neighbors.

\noindent\textbf{Programming model.} We program the NPU
through MLIR-AIR~\cite{mlirair}, a platform-agnostic compiler abstraction 
for spatial accelerators built on MLIR. 
MLIR-AIR defines the AIR dialect, which exposes the array's 
spatial structure as loop nests. AIR models spatial partitioning, 
temporal iteration, and inter-tile communication explicitly, exposing the main deployment decisions for an AI agent to inspect and modify.

IRON~\cite{iron} is another programming abstraction that
sits one layer closer to the hardware. It
exposes compute tiles, memory tiles, and shim tiles directly, with
data movement wired through ObjectFifos and DMA tasks. This
gives expert programmers fine-grained control over tiling,
double buffering, data layout, and pipeline placement, well
suited for hand-tuned kernels.

Dato~\cite{dato} raises the programming model to a task-based dataflow abstraction. It treats on-chip communication and data sharding as first-class program constructs, and expresses an application as a task graph. This shifts low-level tile binding and data movement to the compiler, improving productivity and portability.

\noindent\textbf{LLMs on AMD XDNA NPU.}
AMD Ryzen AI Software~\cite{ryzenai} and
FastFlowLM~\cite{fastflowlm} deploy multiple LLMs on AMD
XDNA NPUs, but they keep their NPU kernel implementation closed-source
 and primarily target quantized LLMs. 
Prior to our work, 
the only end-to-end LLM deployment with open-source
NPU kernels and compiler stack is the BF16 Llama-3.2-1B example
shipped with IRON, which we use as our baseline
in \S\ref{sec:bootstrap}. MLIR-AIR has no prior LLM
deployment example.

\noindent\textbf{Agent skills.} Agent Skills~\cite{anthropicskills} provide a portable format 
for encoding domain knowledge that a coding
agent discovers and invokes autonomously.
Community skill collections such as Superpowers~\cite{superpowers}
apply this format to general software engineering. To our
knowledge, no existing skill set targets end-to-end LLM
deployment on spatial NPUs.

\section{Agent-Assisted Mapping of Llama-3.2-1B}
\label{sec:bootstrap}

This section describes how we mapped a BF16 Llama-3.2-1B end-to-end
onto the AMD XDNA~2 NPU by working with an AI coding agent acting as
our copilot. The deployment outperforms the human-engineered baseline
by 2.2$\times$ on prefill and 4.0$\times$ on decode. We also maintain 
a set of documents that log experiences,
caveats, and the performance optimization trajectory, which
becomes the reusable foundation for the skill system in
\S\ref{sec:skills}.

\subsection{Document-Guided Workflow with a Coding Agent}

Modern AI coding agents such as Claude Code can write code
well, but on a long multi-session project they need a human
to plan and direct, and a persistent way to remember what
has been done. Development on the NPU makes this acute: tooling
documentation is sparse, and opaque error messages surface
frequently. Without a running record of decisions and
debugging, every new session starts blind, easily redoing
past work or breaking past fixes.

To avoid this, we pair a structured development plan with
a set of Markdown documents alongside the code.
The plan has two parts: first reach end-to-end functional
correctness for prefill and decode, then optimize each
path's performance separately. Across both stages,
\texttt{plan.md} captures the overall strategy,
\texttt{progress.md} tracks the current phase, and
\texttt{issues.md} accumulates each non-obvious bug, its
root cause, and the workaround applied. The performance
stage adds \texttt{perf\_opt\_traj.md}, which logs
every optimization attempt with measured before/after
results, alongside per-kernel design notes
(\texttt{gemm.md}, \texttt{attention.md}, \texttt{rope.md},
\ldots) that record each kernel's design rationale,
shape-specific tuning, tile configurations, and known
pitfalls. The human-engineered baseline with IRON serves as
the performance reference at every iteration, both
per-kernel and end-to-end.

Throughout, the coding agent is Claude Code paired with
Claude Opus 4.7 (1M context). The human directs and approves
 while the agent does the search,
code edits, profiling, and tool runs. In each session, the
agent reads the relevant documents to recover state,
proposes the next step against the plan, executes it under
human supervision, and updates the documents as work
progresses. This documentation-first discipline keeps the
workflow durable across sessions.

\subsection{Challenges and the Optimization Trajectory}
\label{sec:trajectory}

\begin{figure}[t]
  \centering
  \includegraphics[width=\columnwidth]{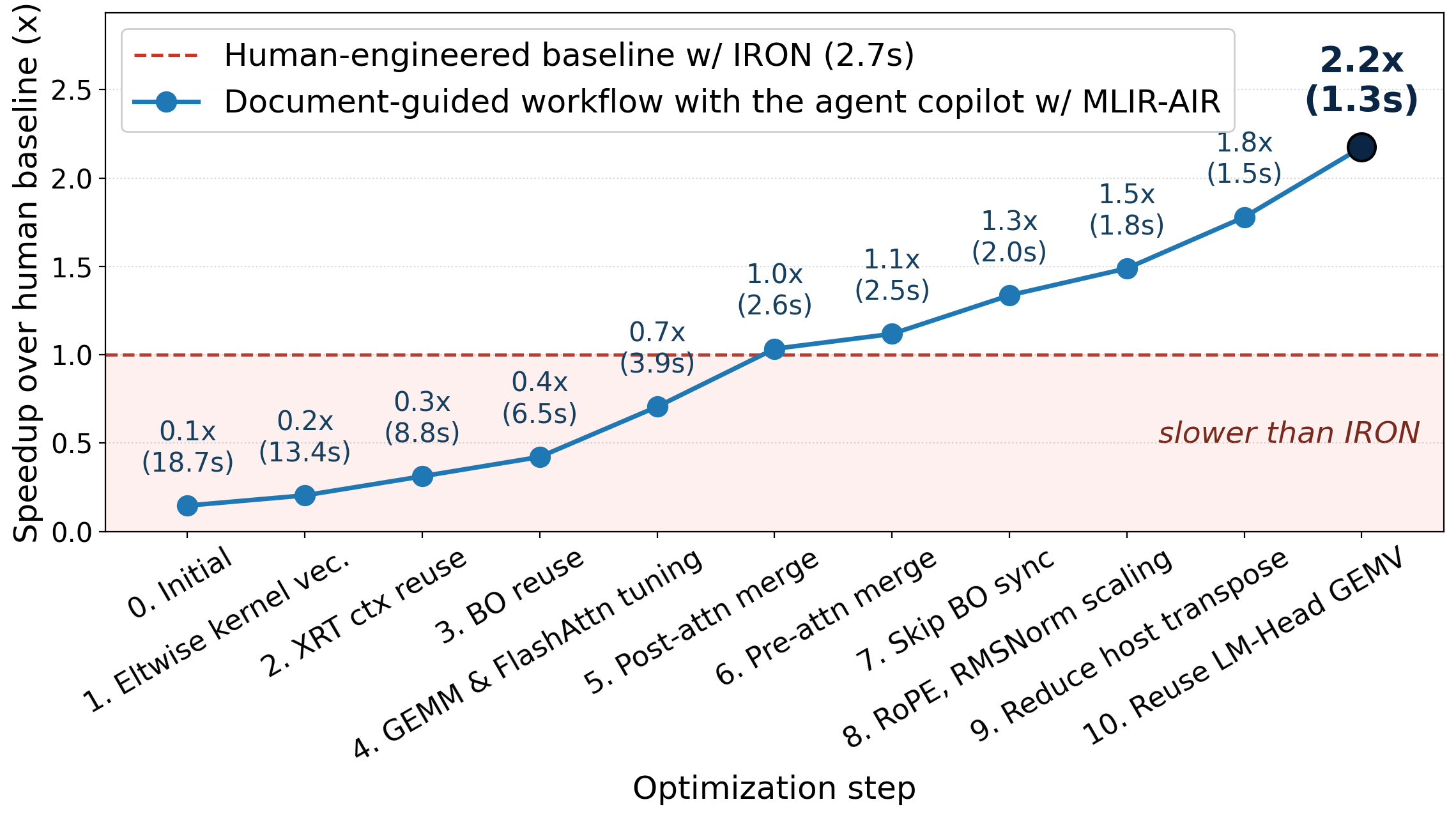}
  \caption{Prefill optimization trajectory of Llama-3.2-1B
    (BF16, seq\_len=2048) on the AMD XDNA~2 NPU.}
  \label{fig:trajectory}
\end{figure}

Mapping the model end-to-end on NPU surfaced challenges in
both correctness and performance. We kept correctness in
check throughout the trajectory with a CPU FP32
reference, gating every kernel and block change against
ground truth. Performance challenges fell into three categories, each
addressed by specific steps in Figure~\ref{fig:trajectory}.

\noindent\textbf{(A) Kernel efficiency at production
shapes.} Reaching peak kernel performance requires
several optimization techniques: vectorization
(Step~1), shape-specific tile tuning (Step~4 GEMM),
fused kernel design (Step~4 FlashAttention), and scaling with
tile-array parallelism (Step~8). 
The main challenge is scaling kernels from validation shapes to production LLM shapes, where implementations must satisfy additional architectural and runtime constraints such as DMA-channel availability, L1/L2 memory capacity, and BufferObject
descriptor limits. 
Cases that expose missing compiler coverage are reported upstream and resolved iteratively.

\noindent\textbf{(B) Reducing kernel dispatch overhead.}
NPU kernel dispatch has non-negligible overhead from the application, runtime, driver, firmware, and hardware layers.
This overhead can exceed the kernel's actual execution time. 
We merged consecutive kernels into single dispatches. 
In prefill, we merged the eight-kernel post-attention block
(Step~5: output projection + SwiGLU FFN) into one dispatch and the six-kernel pre-attention block
(Step~6: RMSNorm + QKV projections + RoPE) into another, reducing per-layer dispatches from 15 to 3. The same merging
applies to decode with GEMV variants. This merging also
saves memory copies of intermediate activations between the
NPU and the host.

\noindent\textbf{(C) Host-side optimization.}
Host-side overheads such as context setup, buffer
management, host-device data transfers, and layout transposes add
up across many kernel calls. We reused the XRT context
across calls (Step~2) and recycled per-layer buffer objects
via zero-copy mapping (Step~3). We skipped redundant host-device
transfers for buffers the NPU writes for static
weights and intermediate activations (Step~7), and chose activation layouts that let
consecutive kernels hand off without host-side transpose
(Step~9). Finally, prefill writes only the last token's logits from
the LM Head, removing a large NPU-to-host transfer of the
full-sequence logits tensor (Step~10).

As a result, on a 2048-token sequence, prefill achieves a
\textbf{2.2$\times$} speedup over IRON (commit 2b62dc7),
with a time-to-first-token (TTFT) of 1.3\,s. Decode achieves
a \textbf{4.0$\times$} speedup, reaching 10.8
tokens/s (TPS). The trajectory and its lessons are logged and distilled
into reusable agent skills in the next section.

\section{Skill System}
\label{sec:skills}

\begin{figure*}[!t]
  \centering
  \includegraphics[width=0.97\textwidth]{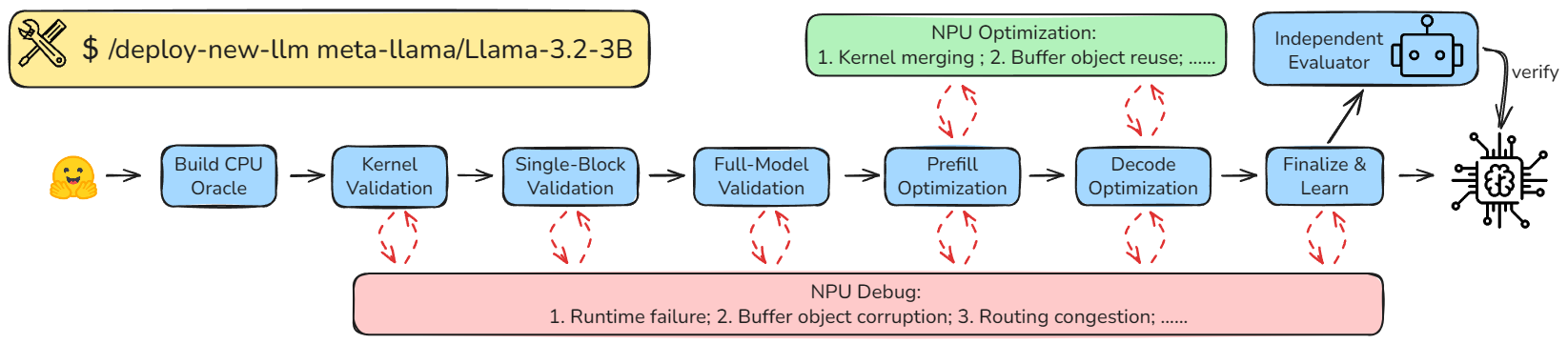}
  \caption{Overview of our skill system for end-to-end LLM deployment on NPU.
  The deploy-new-llm orchestrator dispatches eight phases.
  Phase 4 and 5 draw on optimization skills,
  and any phase can auto-invoke a debug skill on known failures.
  Phases 0-6 each close with a numerical gate against the CPU FP32 reference
  and a human-in-the-loop checkpoint. Phase 7 spawns an independent evaluator
  that re-audits the deployment from scratch.
  }
  \label{fig:overview}
\end{figure*}

Building on the experience of \S\ref{sec:bootstrap},
we distill the maintained document set into a self-evolving
agent skill system to automate the deployment of unencountered LLMs.
Figure~\ref{fig:overview} shows the system and
Table~\ref{tab:skills} catalogs each skill. Each maintained
document maps to a skill component: \texttt{plan.md} and
\texttt{progress.md} define the orchestrator and its phase
skills, the per-kernel design notes (\texttt{gemm.md},
\texttt{attention.md}, \ldots) become a kernel registry used
for Phase~1, \texttt{perf\_opt\_traj.md} becomes the
optimization skills applied in Phase~4 and Phase~5, and
\texttt{issues.md} turns into the auto-invoked debug skills.

\begin{table}[h]
\centering
\caption{All skills in the system, with their type and purpose.}
\label{tab:skills}
\scriptsize
\setlength{\tabcolsep}{2pt}
\begin{tabular}{p{0.30\columnwidth}p{0.16\columnwidth}p{0.46\columnwidth}}
\toprule
\textbf{Skill} & \textbf{Type} & \textbf{Purpose} \\
\midrule
Deploy new LLM & Orchestrator & Scaffold workspace \& dispatch phases \\
\midrule
Build CPU oracle & Phase 0 & Decompose CPU FP32 oracle from HF \\
Kernel validation & Phase 1 & Per-kernel verification vs oracle \\
Single-block validation & Phase 2 & Single transformer block on NPU \\
Full-model validation & Phase 3 & N-layer cascade \& final logits check \\
Prefill optimization & Phase 4 & Apply prefill perf patterns \\
Decode optimization & Phase 5 & Apply decode perf patterns \\
Finalize \& learn & Phase 6 & Runner integration \& lesson harvest \\
Independent evaluator & Phase 7 & Recheck every gate on a fresh subagent \\
\midrule
Multi-kernel merging & Optimization & Merge N kernels into one dispatch \\
Buffer object reuse & Optimization & Load weights with zero-copy mapping \\
Layout alignment & Optimization & Align layouts to skip host transposes \\
\midrule
Runtime failure & Debug & Diagnose kernel hang at runtime \\
Buffer object corruption & Debug & Diagnose BO corruption runtime error \\
Routing congestion & Debug & Diagnose routing congestion hang \\
\bottomrule
\end{tabular}
\end{table}

The eight-phase decomposition mirrors the high-level plan
we followed in \S\ref{sec:bootstrap}. Phases~0--3 establish
end-to-end functional correctness. Phase~0 pulls the
HuggingFace (HF) model and decomposes it into a
kernel-by-kernel CPU FP32 reference verified against the HF
output. Phase~1 verifies each NPU kernel at the model's
required shapes. Phase~2 wires those kernels into a single
transformer block. Phase~3 cascades to the full N-layer
model. Phases~4 and~5 invoke optimization skills
from a shared skill set to address prefill and decode
bottlenecks. Phase~6 integrates prefill and decode into a single
inference runner, profiles the deployment to produce a
performance report, and harvests new findings into
\texttt{lessons.md}. Phase~7 spawns an independent evaluator agent without prior context to re-audit the deployment.

Each of Phases~0--6 closes with a strict numerical gate
against the CPU FP32 reference, measuring absolute error,
relative error, and Pearson correlation across the phase's
output. The gate passes only if all three meet
phase-specific thresholds. For Phases~4 and~5, the gate
additionally requires that each optimization step measurably
improves performance over the prior baseline. If the gate
fails, the agent invokes a matching debug skill on a known
symptom, or asks the user if the failure is a new one. If the gate passes, the agent stops at a human-in-the-loop
checkpoint where the user reviews the phase output and
either approves or redirects. 
Separating Phase~7's evaluator
from the deploying agent prevents reward hacking~\cite{llmjudge}. 
The evaluator re-runs
every numerical gate from scratch and re-profiles to verify that measured performance matches the profiling report submitted by the deploying agent. A fail verdict blocks final acceptance.

Two skill sets complement the phase pipeline: the
optimization skills distilled from \S\ref{sec:trajectory}
patterns, and the debug skills distilled from high-impact bug reports. 
Each set currently contains three entries, and
both are expected to grow gradually. A new debugging fix
or optimization pattern, once verified, becomes a
new entry. At the end of Phase~6, new experiences are
also harvested into \texttt{lessons.md} and fed back into
the skill system for subsequent deployments to inherit.

\section{Evaluation}
\label{sec:eval}

\begin{table*}[ht]
\centering
\caption{End-to-end performance of the deployed LLMs on the AMD XDNA~2 NPU
using our agent skill system.
$\eta_{\text{scale}}$ is the sustained
performance on this model normalized to Llama-3.2-1B's.
Deploy time (h) is the agent's wall clock for entire deployment.
}
\label{tab:deploy}
\footnotesize
\setlength{\tabcolsep}{4pt}
\begin{tabular}{p{0.10\textwidth}rccccccccccccc}
\toprule
\multirow{2}{*}{\textbf{Model}} & \multicolumn{9}{c}{\textbf{Config}} & \multicolumn{2}{c}{\textbf{Prefill}} & \multicolumn{2}{c}{\textbf{Decode}} & \multirow{2}{*}{\textbf{\shortstack{Deploy\\time (h)}}} \\
\cmidrule(lr){2-10} \cmidrule(lr){11-12} \cmidrule(lr){13-14}
 & $\boldsymbol{L}$ & $\boldsymbol{d_{\text{head}}}$ & $\boldsymbol{h/h_{kv}}$ & \textbf{Attn.} & $\boldsymbol{d_{\text{model}}}$ & $\boldsymbol{d_{\text{ffn}}}$ & $\boldsymbol{|V|}$ & \textbf{QKV bias} & \textbf{QK Norm} & \textbf{TTFT (s)} & $\boldsymbol{\eta_{\text{scale}}}$ & \textbf{TPS} & $\boldsymbol{\eta_{\text{scale}}}$ & \\
\midrule
Llama-3.2-1B$^\dagger$ & 16 & 64  & 32/8  & GQA & 2048 & 8192  & 128k & --- & --- & 1.3 & 1.00  & 10.8 & 1.00 & --- \\
\midrule
Llama-3.2-3B  & 28 & 128 & 24/8  & GQA & 3072 & 8192  & 128k & --- & --- & 3.5 & 1.07 & 4.7  & 1.27 & 0.5$^*$ \\
SmolLM2-1.7B  & 24 & 64  & 32/32 & MHA & 2048 & 8192  & 49k  & --- & --- & 2.1 & 1.02 & 7.3  & 1.22 & 0.5$^*$ \\
Qwen2.5-0.5B  & 24 & 64  & 14/2  & GQA & 896  & 4864  & 152k & \checkmark & --- & 0.9 & 0.56 & 8.3  & 0.28 & 4.0$^*$ \\
Qwen2.5-1.5B  & 28 & 128 & 12/2  & GQA & 1536 & 8960  & 152k & \checkmark & --- & 2.6 & 0.67 & 4.9  & 0.60 & 2.5$^*$ \\
Qwen2.5-3B    & 36 & 128 & 16/2  & GQA & 2048 & 11008 & 152k & \checkmark & --- & 3.9 & 0.94 & 4.2  & 1.09 & 1.8 \\
Qwen3-0.6B    & 28 & 128 & 16/8  & GQA & 1024 & 3072  & 152k & --- & \checkmark & 2.3 & 0.24 & 10.5 & 0.48 & 1.5$^*$ \\
Qwen3-1.7B    & 28 & 128 & 16/8  & GQA & 2048 & 6144  & 152k & --- & \checkmark & 2.8 & 0.68 & 6.7  & 0.94 & 1.5$^*$ \\
Qwen3-4B      & 36 & 128 & 32/8  & GQA & 2560 & 9728  & 152k & --- & \checkmark & 8.0 & 0.55 & 2.6  & 0.83 & 2.1 \\
\bottomrule
\end{tabular}

\smallskip
{\footnotesize $^\dagger$ Reference deployment from \S\ref{sec:bootstrap}; 
the rest are deployed autonomously. $^*$ Estimated; the rest are measured from per-phase timing logs.}
\end{table*}

\noindent\textbf{Experiment setup.} All deployments target
the AMD XDNA~2 NPU in Ryzen AI 9 HX 370, programmed via MLIR-AIR. The
coding agent is Claude Code paired with Claude Opus 4.7 at
max reasoning effort. All evaluations
use a sequence length of 2048.

With our skill system, we autonomously deploy eight
additional decoder-only LLMs end-to-end. 
To our knowledge, these are the first open-source end-to-end LLM deployments on the AMD XDNA NPU, 
each completing in 0.5--4 hours with almost no human intervention.
Human input is limited to picking among debug directions the
agent proposes when it hits an unknown bug.

Table~\ref{tab:deploy} reports the configurations and measured
performance of the eight autonomously deployed LLMs alongside
the Llama-3.2-1B reference. The set includes Multi-Head Attention (MHA) and
Grouped-Query Attention (GQA), QKV bias (Qwen2.5 family),
per-head Q/K normalization (Qwen3 family), and tensor shapes
that are not always aligned to tile sizes. All eight LLMs pass
all correctness gates and are also manually verified.

To assess how well the optimizations from our Llama-3.2-1B
reference transferred to the autonomous deployments of
previously unencountered LLMs with different configurations,
we report $\eta_{\text{scale}}$ for each model.
$\eta_{\text{scale}}$ is each new model's sustained
performance relative to Llama-3.2-1B's on the same hardware.
Performance here means achieved compute throughput for
prefill (compute-bound) and achieved memory bandwidth for
decode (memory-bound). $\eta_{\text{scale}}$ captures
hardware utilization rather than absolute latency, so it is
comparable across model sizes.
$\eta_{\text{scale}}=1$ means this model runs the hardware
as efficiently as the reference, $\eta_{\text{scale}}>1$
means it exceeds the reference, and $\eta_{\text{scale}}<1$
means lower efficiency. Full formulas and
calibration are in Appendix~\ref{app:roofline}.

Across the eight deployed models, $\eta_{\text{scale}}$ ranges
0.24--1.07 on prefill and 0.28--1.27 on decode.
Three models (Llama-3.2-3B, SmolLM2-1.7B, Qwen2.5-3B)
reach $\eta_{\text{scale}}$ of 0.94--1.27 across both
prefill and decode. All three share Llama-3.2-1B's basic
transformer block (GQA or MHA over $d_{\text{model}}\geq2048$
with $d_{\text{ffn}}\geq8192$) and differ only in size; the
existing kernels handle them as-is, and their larger GEMMs
amortize the per-launch dispatch overhead identified in
\S\ref{sec:trajectory} more effectively than on the
reference.
The remaining five models reach $\eta_{\text{scale}}$ of
0.24--0.94, which we attribute to two compounding factors.
First, several have small per-kernel work (Qwen3-0.6B has
the smallest $d_{\text{ffn}}$ in our set), so the
per-launch dispatch overhead is not amortized. Second, the
Qwen families introduce ops that the current kernels do
not yet fuse into the projection launches (QKV bias for
Qwen2.5, per-head Q/K normalization for Qwen3), and shapes
not aligned to tile sizes require padding that
wastes useful compute. 
Both are concrete targets for future kernel and agent skill
iterations.

\section{Conclusion and Future Work}

We presented a two-stage methodology for end-to-end LLM
deployment on the AMD XDNA~2 NPU, embodying a transition from
human guidance to agent autonomy. 
Stage one mapped Llama-3.2-1B
with a coding-agent copilot, achieving 2.2$\times$ prefill
and 4.0$\times$ decode speedups over the hand-optimized baseline,
while documenting the trajectory throughout. 
Stage two distilled the documentation into a reusable agent
skill system, which we used to deliver the first open-source
end-to-end deployment of eight additional LLMs
on the AMD NPU.
Each completes in 0.5--4 hours,
with three matching or exceeding the reference's sustained performance.

Several directions remain open. 
New skills can target
additional performance avenues such as kernel fusion,
quantization, and dataflow optimization. 
Architecture coverage can broaden to support sliding-window attention,
Mixture-of-Experts, and Multi-head Latent Attention. 
The methodology can also extend to other spatial
accelerators and coding agents.

\section*{Acknowledgment}
Jiajie Li and Zhiru Zhang were supported in part by NSF Award \#2118709 and a research gift from AMD. 


\clearpage
\bibliographystyle{IEEEtranS}
\bibliography{refs}

\newpage
\appendix
\section{Roofline derivation of $\eta_{\text{scale}}$}
\label{app:roofline}

$\eta_{\text{scale}}$ measures how efficiently a deployed model
uses the NPU relative to our Llama-3.2-1B reference. Higher is
better; $\eta_{\text{scale}}>1$ means that the deployed model exceeds the Llama-3.2-1B reference. We define
\begin{equation}
\eta_{\text{scale}} = \frac{W_{\text{model}} / t_{\text{model}}}{W_{\text{ref}} / t_{\text{ref}}}
\label{eq:eta}
\end{equation}
where $W$ is the dominant work---FLOPs ($F$) for prefill, bytes
($B$) for decode---and $t$ is the measured time. The numerator
is sustained performance (achieved compute throughput for prefill, achieved memory
bandwidth for decode); the denominator normalizes it to the
reference. We estimate $F$ and $B$ below from the model
configuration.

\noindent\textbf{Prefill (compute-bound): $F^{\text{layer}}_{\text{prefill}}$.}
We estimate prefill FLOPs by counting the seven linear
projections (QKV, O, and SwiGLU FFN's gate/up/down) and
FlashAttention; embeddings, RoPE, and norms are negligible and
dropped:
\begin{equation}
F^{\text{layer}}_{\text{prefill}}(S) = 2 S \, d_{\text{model}} (2 d_{\text{model}} + 2 h_{kv} d_{\text{head}} + 3 d_{\text{ffn}}) + 2 S^2 d_{\text{model}}
\label{eq:flops}
\end{equation}
The bracketed term sums the seven projections;
$2 S^2 d_{\text{model}}$ is causal attention. GQA is captured
through $h_{kv}$ (MHA is the special case $h_{kv}=h$).

\noindent\textbf{Decode (memory-bound): $B^{\text{layer}}_{\text{decode}}$.}
At each generated token, the NPU loads the layer's weights from
DRAM (re-loaded per token at batch=1) and reads the KV cache for
the $S$ past tokens. Other transfers---activation vectors, norm
parameters, the new K/V written back---are orders of magnitude
smaller and we drop them. With weights and KV cache both in BF16:
\begin{equation}
\begin{aligned}
B^{\text{layer}}_{\text{decode}}(S) = {} & 2 \, d_{\text{model}} (2 d_{\text{model}} + 2 h_{kv} d_{\text{head}} + 3 d_{\text{ffn}}) \\
                                         & {} + 4 \, S \, h_{kv} d_{\text{head}}
\end{aligned}
\label{eq:bytes}
\end{equation}
The first term is the weight bytes (parameters from
Eq.~\ref{eq:flops}'s bracket, $\times$2 for BF16); the second
is the KV cache (K and V, each $S \times h_{kv} \times d_{\text{head}}$
BF16 elements).

\end{document}